\documentclass[sigconf]{acmart}
\usepackage{algorithmic}
\usepackage{setspace}

\usepackage{multicol}  
\usepackage{multirow}
\usepackage{booktabs}
\usepackage{diagbox}
\usepackage{pifont}
\newcommand{\xmark}{\ding{55}}%
\usepackage{amsmath}
\usepackage{caption}
\usepackage[linesnumbered, ruled]{algorithm2e}
\hypersetup{draft}
\usepackage{enumitem}
\usepackage{bm}
\usepackage{bbm}
\usepackage{tabularx}
\usepackage{graphicx}
\usepackage{subfigure}

\newcolumntype{C}{>{\centering\arraybackslash}X}

\newcommand{\mA}{\mathbf{A}}

\AtBeginDocument{%
  \providecommand\BibTeX{{%
    \normalfont B\kern-0.5em{\scshape i\kern-0.25em b}\kern-0.8em\TeX}}}

\copyrightyear{2023}
\acmYear{2023}
\setcopyright{acmlicensed}\acmConference[CIKM '23]{Proceedings of the 32nd
ACM International Conference on Information and Knowledge
Management}{October 21--25, 2023}{Birmingham, United Kingdom}
\acmBooktitle{Proceedings of the 32nd ACM International Conference on
Information and Knowledge Management (CIKM '23), October 21--25, 2023,
Birmingham, United Kingdom}
\acmPrice{15.00}
\acmDOI{10.1145/3583780.3615250}
\acmISBN{979-8-4007-0124-5/23/10}

\begin{document}
\title{Positive-Unlabeled Node Classification \\
with Structure-aware Graph Learning}



\author{Hansi Yang}
\affiliation{%
  \institution{Department of Computer Science and Engineering \\
  Hong Kong University of Science and Technology}
  \country{Hong Kong, China}}
\email{hyangbw@connect.ust.hk}

\author{Yongqi Zhang}
\affiliation{%
 \institution{4Paradigm}
 \city{Beijing}
 \country{China}}
 \email{yzhangee@connect.ust.hk}

\author{Quanming Yao}
\affiliation{%
 \institution{
 	Department of Electronic Engineering
 	\\
 	Tsinghua University}
 \city{Beijing}
 \country{China}}
\email{qyaoaa@tsinghua.edu.cn}

\author{James Kwok}
\affiliation{%
  \institution{Department of Computer Science and Engineering \\
  Hong Kong University of Science and Technology}
  \country{Hong Kong, China}}
\email{jamesk@cse.ust.hk}

\renewcommand{\shortauthors}{Hansi Yang, Yongqi Zhang, Quanming Yao, \& James Kwok}

\begin{abstract}

Node classification on graphs is an important research problem 
with many applications. 
Real-world graph data sets may not be balanced and accurate as assumed by most existing works. 
A challenging setting is positive-unlabeled (PU) node classification, where labeled nodes are restricted to positive nodes. 
It has diverse applications, e.g., pandemic prediction or network anomaly detection. 
Existing works on PU node classification overlook information in the graph structure, which can be critical. 
In this paper, we propose to better utilize graph structure for PU node classification. 
We first propose a distance-aware PU loss that uses homophily in graphs to introduce more accurate supervision. 
We also propose a regularizer to align the model with graph structure. 
Theoretical analysis shows that minimizing the proposed loss also leads to minimizing the expected loss with both positive and negative labels.
Extensive empirical evaluation on diverse graph data sets demonstrates its superior performance over existing state-of-the-art methods. 
\end{abstract}





\maketitle

\section{Introduction}

Graph-structured data are pervasive in various real-world applications, 
e.g., social networks~\cite{wang2019online}, 
academic networks~\cite{tang2008arnetminer}, 
and traffic networks~\cite{dai2020hybrid}. 
Node classification is a crucial task in analyzing these data 
that can provide valuable insights for their respective domains. 
For instance, on a content-sharing social network (e.g., Flickr), 
content classification enables topic filtering and tag-based retrieval for multimedia items~\cite{seah2018killing}. 
Another example is query graph for e-commerce, 
where classifying queries into different intents can help better focus on the intended product category~\cite{fang2012confidence}. 



\begin{figure}[t]
	\centering
	\includegraphics[width=0.95\linewidth]{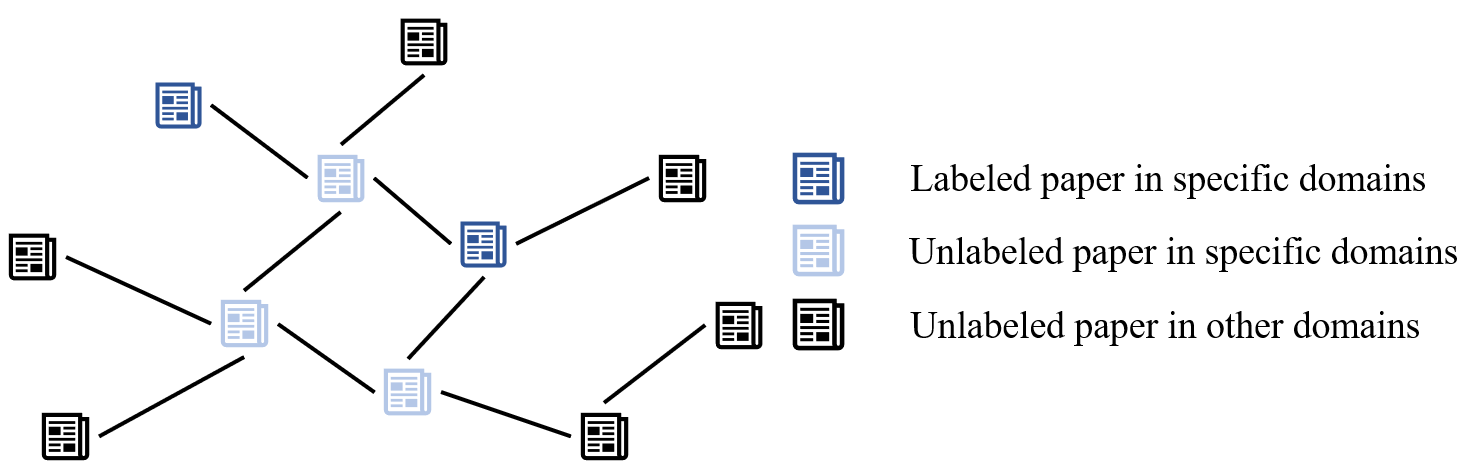}
	\vspace{-10px}
	\caption{An example application for PU node classification: mining papers in specific domains from a citation network.}
	\label{fig:demo}
	\vspace{-15px}
\end{figure}

Existing works on node classification~\cite{velivckovic2017graph,kipf2017semi} 
often assume that labeled nodes are balanced and accurate, 
which may not hold in real world. 
A popular case is positive-unlabeled (PU) learning~\cite{plessis2014upu, kiryo2017nnpu}, 
where we have a binary classification problem and all training data belong to positive class. 
Compared with ordinary semi-supervised learning~\cite{zhou2021semi, berthelot2019mixmatch} for binary classification, 
PU learning is more challenging due to the absence of any known negative samples. 
PU learning also finds natural appearances in graph-structured data, 
especially node-level tasks, 
where we simply restrict all labeled nodes to be positive. 
The above setting, referred as 
PU node classification shown in Figure~\ref{fig:demo}, 
has wide and diverse applications. 
An example is shown in Figure~\ref{fig:demo}, 
where we need to find more papers in specific domains from a citation network. 
Another example is pandemic prediction~\cite{panagopoulos2021transfer}, 
where only positive (affected) nodes are identified, 
and unlabeled nodes can be either positive or negative. 



Previous works on PU learning~\cite{kiryo2017nnpu, zhao2022distpu} 
mainly designed loss functions to effectively learn with only positively-labeled data. 
While these loss functions show better performance than naive baselines (e.g., treating all unlabeled data as negative samples), 
they often assume training samples are independent and identically distributed, 
and cannot leverage graph structures in PU node classification.
Consider the application in Figure~\ref{fig:demo}. 
For an identified positive node (papers known in specific domains), 
its neighboring unlabeled nodes (papers also citing/cited by it) 
are more likely to be positive (i.e., in similar domains)
than those further away. 
However, 
such difference is typically ignored by existing PU learning methods, 
as they do not consider differences among unlabeled samples. 

Motivated by the above limitations, 
in this paper, we propose to make use of graph structures 
for PU node classification. 
We first propose a novel loss function, 
referred as distance-aware PU loss, 
to introduce more accurate supervision. 
We also consider using graph structure as a regularization
to encourage similar representations for neighboring nodes. 
Theoretical analysis demonstrates that 
minimizing our proposed loss 
also leads to minimizing the expected loss with both positive and negative labeled data. 
Empirical results on different graph data sets demonstrate the effectiveness of the proposed method.


\section{Related Work}

\subsection{Learning from Graph-structured Data}




Graph Neural Networks (GNNs) have emerged as a popular approach 
for learning from graph-structured data. 
The key idea is to leverage both node features and graph structures for node representations. 
At each layer of a GNN, 
node representations are updated by aggregating the representations of neighboring nodes. 
After $k$ layers of aggregation, 
the node representations can capture the information of their $k$-hop neighborhoods, 
providing richer information for node classification.
Several GNN architectures have been proposed based on the general idea of neighbor aggregation~\cite{kipf2017semi, velivckovic2017graph, xu2018powerful}. 
For example, Graph Convolutional Network (GCN)~\cite{kipf2017semi} simplifies neighborhood aggregation through graph convolution, similar to the convolution operation for images. 
Graph Attention Network (GAT)~\cite{velivckovic2017graph} applies self-attention mechanism to graph convolution, allowing the model to focus on the most relevant neighbors for each node. 
Graph Isomorphism Network (GIN)~\cite{xu2018powerful} learns more powerful representations from graph structures by incorporating permutation-invariant operations.



\subsection{Positive-Unlabeled Learning} 

Positive-unlabeled (PU) learning refers to 
a special case of binary classification problem
where part of training data is labeled as positive,
and the rest unlabeled data can be either positive or negative.
Existing PU learning methods can be broadly categorized into two approaches: prior-based methods and pseudo-label-based methods.


Prior-based methods assume the knowledge of the class prior, 
i.e., the proportion of positive samples in unlabeled samples, 
and use it to design special loss functions for PU learning. 
Examples include uPU~\cite{plessis2014upu}, nnPU~\cite{kiryo2017nnpu} and Dist-PU~\cite{zhao2022distpu},
which have different forms of loss functions based on different assumptions.  
Other works, sush as~\cite{chen2020selfpu, gong2022instance}, introduce other types of supervision to PU learning, 
e.g., model distillation with self-paced curriculum in~\cite{chen2020selfpu}. 

In contrast,
pseudo-label-based methods use two heuristic steps: 
first, identifying reliable negative samples from the unlabeled data,
and then performing (semi-)supervised learning with additional pseudo labels. 
For example, PUbN~\cite{hsieh2019pubn} uses a model pretrained by nnPU to recognize high-confidence negative samples in the unlabeled data. 
PULNS~\cite{luo2021pulns} incorporates reinforcement learning to obtain an effective negative sample selector.

\section{Methodology}
\label{sec:method}
For
a graph
$\mathcal{G}=(\mathcal{V},\mathcal{E}, \mathbf{X})$,
we use $\mathcal{V}=\{v_1,\dots,v_N\}$ to denote the set of $N$ nodes, 
$\mathcal{E} \subseteq \mathcal{V} \times \mathcal{V}$ to denote the set of edges,
and $\mathbf{X}=\{\mathbf{x}_1,\dots,\mathbf{x}_N\}$ to denote the set of node attributes, 
with $\mathbf{x}_i$ being the attributes for node $v_i$. 
Denote the adjacency matrix
by $\mathbf{A} \in \mathbb{R}^{N \times N}$,
where $\mathbf{A}_{ij}=1$ if nodes ${v}_i$ and ${v}_j$ are connected, and 0
otherwise. 
Let $\mathcal{V}_L$ be
the set of labeled positive nodes, and $\mathcal{V}_U = \mathcal{V}-\mathcal{V}_L$ be the set of unlabeled nodes. 
With these notations, 
the 
problem  of
PU node classification 
can be defined as follows.
\newtheorem{problem}{Problem}
\begin{problem}
Given a graph $\mathcal{G}=(\mathcal{V}, \mathcal{E}, \mathbf{X})$ with a 
set of positively-labeled nodes $\mathcal{V}_L$, 
we want to learn a model
 $f$ (a GNN here) which predicts the true labels 
$\hat{\mathcal{Y}}_U$ 
of the unlabeled nodes $\mathcal{V}_U$,
i.e.,
\begin{equation*}
f(\mathcal{G}, \mathcal{V}_L, \mathcal{V}_U) \rightarrow \mathcal{\hat{Y}}_U.
\end{equation*}
\end{problem}
An example is illustrated in Figure~\ref{fig:demo}, 
which aims to find papers in specific domains from a citation network. 
$\mathcal{V}_L$ contains papers that are known in these domains,
while $\mathcal{V}_U$ contains papers whose domains are unknown. 
Compared to standard PU learning, 
a critical difference of PU node classification 
is the graph structure.
In particular, 
homophily, i.e.,
connected nodes are more likely to belong to the same class, 
has been observed in 
many 
graph
applications.
For instance, in an academic network~\cite{tang2008arnetminer}, 
papers are more likely to cite papers in the same field. 
In an epidemic network~\cite{panagopoulos2021transfer}, 
neighbors of positive (affected) nodes are also more likely to be positive. 
Nevertheless, such information has been ignored by previous works on PU node classification~\cite{wu_learning_2021, yoo2021accurate}, 
which simply use the general PU loss for node classification. 

To address this problem, in this section
we consider leveraging the graph structure to provide more supervision for model
training.  
Specifically, we propose the positive-unlabeled GNN (PU-GNN) 
with a distance-aware PU loss (section~\ref{sec:stuloss})
as well as a regularizer from graph structure (section~\ref{ssec:reg})
to 
help model training. 




\subsection{Distance-aware PU Loss}
\label{sec:stuloss}


There have been many works on designing special loss functions for PU learning~\cite{plessis2014upu, kiryo2017nnpu, zhao2022distpu}. 
One of the state-of-the-art work, Dist-PU~\cite{zhao2022distpu}, 
defines the loss as follows: 
\begin{align}
\mathcal{L}_{\text{Dist-PU}} 
\! = \! 
2 \pi_P 
\left| \frac{1}{n_L} \! \sum\nolimits_{v_i \in \mathcal{V}_L} \hat{y}_i \! - \! 1 \right| 
+ 
\left| \frac{1}{n_U} \! \sum\nolimits_{v_i \in \mathcal{V}_U} \hat{y}_i \! - \! \pi_P \right|,
\label{eq:distpu}
\end{align}
where $n_L=|\mathcal V_L|$ denotes the number of labeled nodes, 
$n_U=|\mathcal V_U|$ denotes the number of unlabeled nodes, 
and $\pi_P$ is the prior probability for the positive class. 
A critical shortcoming of existing works like \eqref{eq:distpu} is the ignorance of structural information for PU learning on graph. 
Therefore, in the following, we introduce how to integrate graph structure into the design of loss functions. 

Consider a graph with only one positive node,
while all
other nodes are unlabeled.
From homophily, the connected nodes 
are more likely to belong to the same class, 
and the influence of the only positive node should decrease with 
its distance to other nodes. 
As such, the distance to this positive node can help determine the node label: 
the closer an unlabeled node is to the positive node,
the more likely it is to belong to the positive class. 


When the graph has multiple positive nodes,
a simple extension is to consider the distance of each unlabeled
node to its nearest positive node.
Mathematically, 
let $\text{dist}(v, v')$ be the shortest-path distance between two nodes $v$ and $v'$.
We first propose to split all the unlabeled nodes in $\mathcal{V}_U$ into two\footnote{Preliminary results show that using more subsets does not lead to better empirical performance.}
subsets 
$\hat{\mathcal{V}}_U$
and
$\breve{\mathcal{V}}_U$: 
\begin{align*}
\hat{\mathcal{V}}_U = 
\Big\{ v: v\in \mathcal{V}_U, \min\nolimits_{v' \in  \mathcal{V}_L} \text{dist}(v, v') \le \delta \Big\}, \quad\breve{\mathcal{V}}_U = \mathcal{V}_U \setminus \hat{\mathcal{V}}_U,
\end{align*}
where 
$\delta$ is a  given threshold.
Based on the above two subsets, we propose the  following loss
function for PU node classification:
\begin{align} 
\small
\mathcal{L}_\mathcal{G} = 
& 2\Big( \hat{\pi}_P + \breve{\pi}_P \Big)
\Big| \frac{1}{n_L} \!\sum\nolimits_{v_i \in \mathcal{V}_L} \!\hat{y}_i \!-\! 1 \Big| 
\nonumber
\\ 
& + \Big| \frac{1}{\hat{n}_U} \!\sum\nolimits_{v_i \in \hat{\mathcal{V}}_U}\!\hat{y}_i \!-\! \hat{\pi}_P \Big| 
+ \Big| \frac{1}{\breve{n}_U} \!\sum\nolimits_{v_i \in \breve{\mathcal{V}}_U} \!\hat{y}_i 
\!-\! \breve{\pi}_P \Big|,
\label{eq:str}
\end{align}
where $n_L=|\mathcal V_L|$, $\hat{n}_U = |\hat{\mathcal{V}}_U|, \breve{n}_U = |\breve{\mathcal{V}}_U|$,
$\hat{\pi}_P$ (resp. $\breve{\pi}_P$)
is the 
proportion of positive 
samples in $\hat{\mathcal{V}}_U$ (resp. $\breve{\mathcal{V}}_U$),
and $\hat{y}_i$ is the prediction score for node $i$. 
Naturally, the choice of $\hat{\pi}_P$ and $\breve{\pi}_P$ should satisfy $\hat{\pi}_P > \breve{\pi}_P$, 
as nodes closer to labeled positive nodes should 
have higher probability of being positive. 
We note that each term in (\ref{eq:str}) corresponds to a set of nodes ($\mathcal{V}_L$, $\hat{\mathcal{V}}_U$ and $\breve{\mathcal{V}}_U$). 
With only one unlabeled data set 
(i.e., not using structural information), 
the loss in
(\ref{eq:str})
reduces to the Dist-PU loss in (\ref{eq:distpu}). 
Similar to~\cite{zhao2022distpu}, the following proposition shows that the
proposed loss provides an upper bound on the training loss with ground-truth labels (both positive and negative).
\begin{proposition}
Denote the distribution of positive (resp. negative) nodes 
as $p_P(v)$ (resp. $p_N(v)$). 
For each node $v_i$, denote its prediction as $y_i$, 
and define the expected loss on the whole graph as 
$\hat{\mathcal{L}} = \pi_P | 
\mathbb{E}_{v_i \sim p_P(v_i)} \hat{y}_i-1 | 
+ ( 1-\pi_P ) 
| \mathbb{E}_{v_i \sim p_N(v_i)} \hat{y}_i |$. 
Then for a class of bounded 
functions $F$ with VC dimension $V$, with probability at least 1 - $\delta$, we
have:
\begin{align*}
\small 
\hat{\mathcal{L}} \le \mathcal{L}_\mathcal{G} & + 8\pi_P C\sqrt{V / n_L} 
+ 12 \pi_P \sqrt{ \log(4/\delta) / (2 n_L) } \\
& + 4C \sqrt{V/(\hat{n}_U + \breve{n}_U)} + 6 \sqrt{ \log(4/\delta) / (2\hat{n}_U + 2\breve{n}_U)},
\end{align*}
\end{proposition}


The above proposition demonstrates that 
minimizing $\mathcal{L}_\mathcal{G}$ also leads to minimizing 
the expected loss $\hat{\mathcal{L}}$ by minimizing the upper bound on the right hand side, 
though we do not know any negatively-labeled nodes in practice. 

\subsection{Structural Regularization}
\label{ssec:reg}

While distance-aware PU loss assigns different priors among unlabeled nodes, 
it still overlooks the pairwise relations between nodes. 
As such, 
we propose a regularization term based on the graph structure 
to encourage similar representations and final predictions for 
neighboring nodes. 

Let $P(v_i)$ be the uniform distribution on all nodes that are \textit{not} node $v_i$'s neighbors. 
For each edge
$(i,j)$ with $\mA_{ij}=1$, 
we randomly sample $K$ nodes 
from 
$P(v_i)$
as negative samples.
The structural regularizer $\mathcal{R}_S$ is then defined as:
\begin{equation}
\small
\mathcal{R}_S
\! = \! \sum\nolimits_{v_i \in \mathcal{V}}
\sum\nolimits_{v_j \in \mathcal{N}(v_i)} 
\Big((\mathbf{S}_{ij} \! - \! 1)^2 
\! + \! 
\sum\nolimits_{k=1}^K \mathbb{E}_{v_k \sim P(v_i)} (\mathbf{S}_{ik} \! - \! 0)^2\Big),
\label{eq:edge}
\end{equation}
where $\mathbf{S}_{ij}=\sigma(\mathbf{z}_i^{\top} \mathbf{z}_j)$ is the similarity between node $v_i$ and $v_j$'s representations
$\mathbf{z}_i$ and $\mathbf{z}_j$, 
$\sigma(\cdot)$ is the sigmoid function,
and $\mathcal{N}(v_i)$ is the set of node $v_i$'s neighboring nodes.
Minimizing the first term $(\mathbf{S}_{ij} \! - \! 1)^2$ 
in (\ref{eq:edge})
encourages neighboring nodes ($v_i$ and $v_j$) to have similar representations, 
while minimizing the second term $\sum\nolimits_{k=1}^K \mathbb{E}_{v_k \sim P_n(v_i)} (\mathbf{S}_{ik} \! - \! 0)^2$ 
in (\ref{eq:edge})
encourages node $v_i$ to have different representations from its
non-neighboring
nodes $v_k$'s.

Combining \eqref{eq:edge} and \eqref{eq:str}, the final objective $\mathcal{L}$ is given by 
$\mathcal{L} = \mathcal{L}_\mathcal{G} +  \alpha \mathcal{R}_S$,
where $\alpha$ is used to balance the contribution of 
structural regularization $\mathcal{R}_S$.

\section{Experiments}
\label{Sec:experiments}



\subsection{Experimental Setup}
\label{Sec:ex_settings}



We conducted experiments on
 four node classification data sets: 
\textit{Cora}, \textit{Citeseer}, \textit{Pubmed} and \textit{DBLP}, 
which are popularly used in literature~\cite{kipf2017semi, velivckovic2017graph, wu_learning_2021, yoo2021accurate}.
The data set statistics are summarized in Table~\ref{tab:dataset}.
We follow the splits in \cite{sen2008collective},
but merge their validation and test splits into a single test split. 
To transform these datasets into binary classification tasks, we follow the approach in~\cite{wu_learning_2021},
where we treated a subset of the classes as positive and the others as negative. 
The resulting numbers of positive and negative samples are shown in Table~\ref{tab:dataset}.
To convert these datasets into PU learning problems, 
we randomly selected a small proportion of positive nodes (called {\em label ratio}) 
in the training set and removed the labels of the rest.
We experimented with different label ratios,
including $\{0.001, 0.002, 0.005, 0.01\}$, 
to evaluate the performance of our approach under varying levels of label scarcity.


\begin{table}[H]
\vspace{-10px}
	\caption{Statistics of datasets.}
	\vspace{-10px}
	\centering
\begin{tabular}{cc|cccc}
		\hline
&		& \textit{Cora} & \textit{Citeseer} & \textit{Pubmed} & \textit{DBLP} \\
		\hline
\# nodes & total & 2,708 & 3,327 & 19,717 & 17,716 \\
& positive & 1,244 & 1,369 & 7,875 & 7,920 \\
& negative & 1,464 & 1,958 & 11,842 & 9,796\\
& training & 271 & 333 & 1,971 & 1,772 \\
& testing & 2,437 & 2,994 & 17,746 & 15,944 \\
		\hline
	\end{tabular}
	\label{tab:dataset}
	\vspace{-10px}
\end{table}

For the proposed method,
we set $\alpha=0.01$, threshold $\delta=3$, 
the number of negative samples $K=50$,  
and the class priors $\hat{\pi}_P = 0.6, \breve{\pi}_P=0.3$.
We use a two-layer GCN with hidden dimension 16 as the backbone.
As will be demonstrated by experiments later, 
the proposed method is not sensitive to these hyper-parameters
and it is easy to pick up proper values. 

The proposed method 
is compared 
with the following baselines:
(i) naive GCN, which considers all unlabeled nodes as negative;
(ii) GCN with nnPU loss~\cite{kiryo2017nnpu};
(iii) GCN with Dist-PU loss~\cite{zhao2022distpu}; 
(iv) LSDAN~\cite{wu_learning_2021},
which introduces long-range dependency with the nnPU loss~\cite{kiryo2017nnpu}; 
and (v) GRAB~\cite{yoo2021accurate}, 
which models the given graph as a Markov network 
and uses loopy belief propagation (LBP) 
to estimate the label distribution of unlabeled nodes. 
As in~\cite{kipf2017semi, kiryo2017nnpu, zhao2022distpu},
we use 
macro F1 score (\%) on the testing nodes
as performance measure.
The experiment is repeated
5 times, and we report 
the average performance with standard deviation.

\subsection{Node Classification Performance}


Table~\ref{tab:results} shows
macro F1 scores (\%) on the testing set
with different label ratios.
As can be seen,
using the PU loss improves the naive GCN baseline much, 
and Dist-PU significantly outperforms nnPU in most cases. 
Methods specific to PU node classification (LSDAN~\cite{wu_learning_2021} and GRAB~\cite{yoo2021accurate}) 
generally have better performances than Dist-PU, 
and their performances are comparable to each other. 
The proposed PU-GNN achieves the best performance for almost all
data sets and label ratios,
demonstrating that graph structure
plays a critical role in the loss design for PU node classification. 

\begin{table}[h]
    \centering
    \vspace{-10px}
    \caption{Macro F1 score (\%) on the testing set with different label ratio $\{0.001, 0.002, 0.005, 0.01\}$. 
   	GCN is used as the backbone network for all methods.} 
	\vspace{-10px}
	\setlength{\tabcolsep}{3.5px}
    \begin{tabular}{c c | c c c c }
    \hline
     Data set & Method & 0.001 & 0.002 & 0.005 & 0.01 \\
    \hline
    \multirow{6}{*}{\textit{Cora}} & naive & 78.1$\pm$0.1 & 78.1$\pm$0.1 & 78.1$\pm$0.1 & 78.1$\pm$0.1 \\
    & nnPU & 80.3$\pm$0.2 & 80.9$\pm$0.3 & 83.9$\pm$0.4 & 85.6$\pm$0.2 \\
    & Dist-PU & 82.1$\pm$0.2 & 82.7$\pm$0.3 & 86.6$\pm$0.3 & 87.4$\pm$0.2 \\
    & LSDAN & 81.8$\pm$0.2 & 82.8$\pm$0.2 & 86.7$\pm$0.3 & 87.6$\pm$0.2 \\
    & GRAB & 82.5$\pm$0.3 & 84.4$\pm$0.3 & 86.5$\pm$0.2 & 87.5$\pm$0.2 \\
    & PU-GNN & \textbf{84.8$\pm$0.3} & \textbf{86.0$\pm$0.3} & \textbf{87.3$\pm$0.2} & \textbf{88.3$\pm$0.2} \\
    \hline
    \multirow{6}{*}{\textit{Pubmed}} & naive & 57.1$\pm$0.1 & 57.1$\pm$0.1 & 57.1$\pm$0.1 & 57.1$\pm$0.1 \\
    & nnPU & 63.6$\pm$0.4 & 65.7$\pm$0.3 & 70.5$\pm$0.3 & 73.8$\pm$0.3 \\
    & Dist-PU & 63.3$\pm$0.3 & 67.1$\pm$0.2 & 72.2$\pm$0.2 & 74.3$\pm$0.3 \\
    & LSDAN & 69.6$\pm$0.4 & 71.4$\pm$0.3 & 72.7$\pm$0.2 & \textbf{75.1$\pm$0.2} \\
    & GRAB & 70.6$\pm$0.3 & 71.6$\pm$0.4 & 73.3$\pm$0.2 & 75.0$\pm$0.3 \\
    & PU-GNN & \textbf{73.0$\pm$0.4} & \textbf{73.5$\pm$0.3} & \textbf{74.0$\pm$0.2} & 74.8$\pm$0.2\\
    \hline
    \multirow{6}{*}{\textit{Citeseer}} & naive & 61.0$\pm$0.1 & 61.0$\pm$0.1 & 61.0$\pm$0.1 & 61.0$\pm$0.1 \\
    & nnPU & 61.2$\pm$0.4 & 62.2$\pm$0.3 & 62.7$\pm$0.2 & 63.8$\pm$0.2 \\
    & Dist-PU & 61.4$\pm$0.4 & 62.2$\pm$0.2 & 63.5$\pm$0.3 & 64.6$\pm$0.2 \\
    & LSDAN & 62.7$\pm$0.3 & 63.6$\pm$0.2 & 65.6$\pm$0.3 & 70.5$\pm$0.3 \\
    & GRAB & 61.8$\pm$0.4 & 62.6$\pm$0.3 & 65.2$\pm$0.3 & 69.7$\pm$0.4 \\
    & PU-GNN & \textbf{64.5$\pm$0.4} & \textbf{64.7$\pm$0.2} & \textbf{65.8$\pm$0.3} & \textbf{73.0$\pm$0.3} \\
    \hline
    \multirow{6}{*}{\textit{DBLP}} & naive & 34.4$\pm$0.2 & 34.4$\pm$0.2 & 34.4$\pm$0.2 & 34.6$\pm$0.3 \\
    & nnPU & 83.4$\pm$0.3 & 83.9$\pm$0.2 & 85.4$\pm$0.3 & 86.7$\pm$0.4 \\
    & Dist-PU & 83.8$\pm$0.2 & 84.1$\pm$0.3 & 86.8$\pm$0.2 & 87.3$\pm$0.3 \\
    & LSDAN & 85.3$\pm$0.2 & 85.6$\pm$0.2 & 86.8$\pm$0.2 & \textbf{87.8$\pm$0.2} \\
    & GRAB & 85.0$\pm$0.2 & 85.7$\pm$0.3 & 86.8$\pm$0.2 & \textbf{87.8$\pm$0.2} \\
    & PU-GNN & \textbf{86.5$\pm$0.1} & \textbf{86.6$\pm$0.2} & \textbf{87.2$\pm$0.4} & \textbf{87.8$\pm$0.3} \\
    \hline
    \end{tabular}
    \vspace{-10px}
    \label{tab:results}
\end{table}

\subsection{Ablation Study}

\subsubsection{Contribution of Each Component}
\label{sec:ablation}
We first study the usefulness of the
proposed distance-aware PU loss (abbreviated as ``PU loss'') and structural regularization (abbreviated as ``struct reg''). 
Table~\ref{tab:abl} shows macro F1 scores (\%) on the testing set of {\it Cora} 
for GCN and different variants of
the proposed PU-GNN
with different label ratios.
As can be seen, both the distance-aware PU loss and structural regularization
are critical to model performance. 
Moreover, 
with more labels, using only distance-aware PU loss achieves closer performance
with the full PU-GNN model, 
and even slightly outperforms the full PU-GNN model at the end. 
This can be explained 
as more labeled nodes provide more accurate supervision,  
leading to better performances even when
only the distance-aware PU loss
is used.

\begin{table}[t]
    \centering
    \vspace{-10px}
    \caption{Macro F1 score (\%) on the testing set with different variants of PU-GNN. 
	 \textit{Cora} data set is used.}
    \vspace{-10px}
    \begin{tabular}{c | c| cc}
    \hline
Label     ratio & GCN & \multicolumn{2}{c}{PU-GNN} \\
    \hline
& & ($\checkmark$ struct reg, $\checkmark$ PU loss) &  \textbf{84.8$\pm$0.3}  
\\\cline{3-4}
0.001 & 78.1$\pm$0.1 &
($\checkmark$ struct reg, \xmark\; PU loss) &  80.2$\pm$0.2 \\\cline{3-4}
&& (\xmark\; struct reg, $\checkmark$ PU loss) &  83.2$\pm$0.3  \\\hline
& & ($\checkmark$ struct reg, $\checkmark$ PU loss) &  \textbf{86.0$\pm$0.3}  
\\\cline{3-4}
0.002 & 78.1$\pm$0.1 &
($\checkmark$ struct reg, \xmark\; PU loss) &  80.6$\pm$0.3 \\\cline{3-4}
&& (\xmark\; struct reg, $\checkmark$ PU loss) &  85.0$\pm$0.4 \\\hline
& & ($\checkmark$ struct reg, $\checkmark$ PU loss) &  \textbf{87.3$\pm$0.2}  
\\\cline{3-4}
0.005 & 78.1$\pm$0.1 &
($\checkmark$ struct reg, \xmark\; PU loss) & 81.6$\pm$0.2 \\\cline{3-4}
&& (\xmark\; struct reg, $\checkmark$ PU loss) &  86.7$\pm$0.3 \\\hline
& & ($\checkmark$ struct reg, $\checkmark$ PU loss) &  88.3$\pm$0.2
\\\cline{3-4}
0.01 & 78.1$\pm$0.1 &
($\checkmark$ struct reg, \xmark\; PU loss) & 83.4$\pm$0.3 \\\cline{3-4}
&& (\xmark\; struct reg, $\checkmark$ PU loss) &  \textbf{88.7$\pm$0.3}  \\\hline
    \end{tabular}
    \vspace{-15px}
    \label{tab:abl}
\end{table}



\subsubsection{Hyper-parameter Sensitivity}
Here, we study the proposed method's sensitivity to hyper-parameters. 
We use the \textit{Cora} data set with label ratio 0.002.
For $\hat{\pi}_P$ and $\breve{\pi}_P$, 
we should have $\hat{\pi}_P, \breve{\pi}_P \in (0,1)$ and $\hat{\pi}_P \ge \breve{\pi}_P$, 
since they refer to the proportion of positive samples in a given node set.
And for the threshold $\delta$ in \eqref{eq:str}, we vary it in $\{1,2,3,4,5\}$.  
We do not consider $\delta \ge 6$ as for all the three data sets, 
the largest distance between any pairs of nodes is not larger than $6$, 
hence using $\delta \ge 6$ will make $\breve{\mathcal{V}}_U$ be an empty set and reduce to the Dist-PU loss. 

Table~\ref{tab:pi} shows macro F1 scores (\%) on the testing set 
with different positive priors. 
We can see that our method attains satisfying performance across a wide range of values around $\hat{\pi}_P=0.6$ and $\breve{\pi}_P = 0.3$, 
where performances better than any baseline methods in Table~\ref{tab:results} are all highlighted. 

\begin{table}[h]
\vspace{-10px}
    \caption{Macro F1 score (\%) on the testing set with different $\hat{\pi}_P$ and $\breve{\pi}_P$'s. \textit{Cora} data set is used.}
    \vspace{-10px}
    \centering    
    \setlength{\tabcolsep}{4.5px}
    \begin{tabular}{c | c c c c c}
    \hline
    \diagbox{$\breve{\pi}_P$}{$\hat{\pi}_P$} & 0.5 & 0.6 & 0.7 & 0.8 & 0.9 \\
    \hline
     0.1 & \textbf{84.6$\pm$0.3} & \textbf{85.3$\pm$0.2} & \textbf{84.6$\pm$0.2} & 83.8$\pm$0.4 & 83.4$\pm$0.3 \\
     0.2 & \textbf{84.6$\pm$0.3} & \textbf{85.3$\pm$0.2} & \textbf{85.2$\pm$0.3} & 84.0$\pm$0.3 & 83.4$\pm$0.3 \\
     0.3 & \textbf{85.7$\pm$0.3} &  \textbf{86.0$\pm$0.3} & \textbf{85.9$\pm$0.3} & 84.8$\pm$0.3 & 84.4$\pm$0.2 \\
     0.4 & 84.0$\pm$0.3 & \textbf{85.9$\pm$0.3} & 83.7$\pm$0.4 & 84.0$\pm$0.4 & 83.4$\pm$0.3 \\
    \hline
    \end{tabular}
    \label{tab:pi}
    \vspace{-10px}
\end{table}

Table~\ref{tab:delta} compares the macro F1 scores (\%) on the testing set 
with different thresholds $\delta$.
As can be seen, setting $\delta$ too small (resp. too large) will make set $\hat{\mathcal{V}}_U$ (resp. $\breve{\mathcal{V}}_U$) 
only have a small amount of nodes, 
leading to similar performances with Dist-PU loss (82.7$\pm$0.3). 
The proposed method achieves good performances when $\delta$ is chosen from 2 or 3. 


\begin{table}[h]
\vspace{-8px}
    \caption{Macro F1 score (\%) on the testing set with different $\delta$'s. \textit{Cora} data set is used. }
    \vspace{-10px}
    \centering
    \setlength{\tabcolsep}{3px}
    \begin{tabular}{c | c c c c c c}
    \hline
    $\delta$ & 1 & 2 & 3 & 4 & 5 \\
    \hline
    F1 score (\%) & 83.0$\pm$0.4 & \textbf{85.8$\pm$0.2} & \textbf{86.0$\pm$0.3} & 83.2$\pm$0.3 & 82.8$\pm$0.3 \\
    \hline
    \end{tabular}
    \label{tab:delta}
    \vspace{-10px}
\end{table}


%


\section{Conclusion}

In this paper, 
we propose to utilize graph structures to benefit PU node classification. 
We first propose a distance-aware loss function, 
which uses homophily in graph structure to introduce more accurate supervision for unlabeled nodes. 
Theoretical analysis demonstrates that 
minimizing the proposed loss leads to minimizing the expected loss with both positive and negative labels. 
We also propose a regularizer based on graph structure 
to further improve model performances. 
Empirical results across different data sets demonstrate the effectiveness of our proposed method. 
For future works, we may consider generalize the proposed method from homophily to heterophily for node classification. 

\section*{Acknowledgment}
This research was supported in part by the Research Grants Council of the Hong Kong Special Administrative Region (Grant 16200021).
Q. Yao is supported by NSF of China (No. 92270106).



%

\clearpage
\bibliographystyle{ACM-Reference-Format}
\bibliography{acmart}

\end{document}